\documentclass[11pt,letterpaper]{article}
\usepackage[T1]{fontenc}
\usepackage{lmodern}
\usepackage{microtype}
\usepackage[margin=1in]{geometry}
\usepackage{amsmath}
\usepackage{booktabs}
\usepackage{caption}
\usepackage{graphicx}
\usepackage{standalone}
\usepackage{tikz}
\usetikzlibrary{arrows.meta}
\usepackage{authblk}

\usepackage[colorlinks,allcolors=blue!50!black]{hyperref}
\newcommand{\promptbox}[2]{%
    \noindent
    \begingroup
    \setlength{\fboxsep}{6pt}%
    \fcolorbox{black!30}{black!4}{%
        \begin{minipage}[t]{\dimexpr\linewidth-2\fboxsep-2\fboxrule\relax}
            \small\raggedright
            \textbf{#1}\par\smallskip
            {\ttfamily #2\par}
        \end{minipage}}%
    \endgroup
}

\title{Multi-Resolution Parallel Scaling for Video Question Answering at TREC 2026}

\author[1,2]{Andreas Goulas}
\author[2]{Ioannis Patras}
\author[1]{Vasileios Mezaris}

\affil[1]{CERTH-ITI, Thessaloniki, Greece}
\affil[ ]{\texttt{\{agoulas, bmezaris\}@iti.gr}}
\affil[2]{Queen Mary University of London, London, UK}
\affil[ ]{\texttt{i.patras@qmul.ac.uk}}

\date{}

\begin{document}

\maketitle

\begin{abstract}
Parallel inference-time scaling refers to a group of large language model (LLM) scaling techniques that increase inference-time compute via multiple parallel generations.
In this paper, we present a novel parallel scaling approach for Video LLMs with reasoning, developed for the CERTH-ITI participation in the TREC 2026 Video Question Answering (VQA) track.
We address both multiple-choice (MC) and open-ended answer generation (AG) video QA tasks, which require spatio-temporal reasoning and audio-visual understanding.
In contrast to prior work that uses a fixed spatio-temporal resolution across all parallel generation streams, our proposed approach observes the input video under varying spatio-temporal resolutions to capture complementary aspects of the input signal.
Information is shared between the $P$ parallel streams by averaging the logits at each decoding step.
Motivated by self-consistency decoding, we perform up to $D$ sequential generation attempts, each conditioned on $P$ jointly processed streams, and rank the accepted answers by frequency.
For the AG task, we perform training-free adaptation via few-shot prompting by independently sampling examples for each stream and generation attempt, further increasing the diversity of the parallel generations.
We report and analyze the results of 4 AG and 5 MC runs submitted to TREC 2026, observing performance gains for the main two-stream parallel configurations over the single-stream Chain-of-Thought baselines.
Finally, we report approximate run durations and recorded generation times per submitted answer, observing a modest latency impact when scaling to $P=2$ parallel generations.
\end{abstract}

\section{Introduction}

Scaling LLMs by increasing the number of parallel generations is a well-established test-time scaling approach that is typically applied in combination with standard Chain-of-Thought (CoT) reasoning \cite{wei2022chain} to aggregate information from diverse reasoning trajectories.
Parallel scaling offers an alternative to extending individual reasoning sequences, which can suffer from overthinking \cite{gema2025inverse}.
Furthermore, parallel scaling has been shown to have a low to modest impact on generation latency for small batches, due to the unique characteristics of LLM inference \cite{chen2026parallel}.
In the video domain, recent work has adapted the parallel scaling paradigm by designing frameworks where different streams observe diverse frame subsets of the input video \cite{goulasmllm,suo2025trial,chung2026video}.
This has the effect of expanding the visual coverage of the model.
However, these works typically use a fixed spatio-temporal resolution for all streams, limiting the diversity of the parallel representations.

\begin{figure}[tbp]
    \centering
    \includegraphics[width=\linewidth]{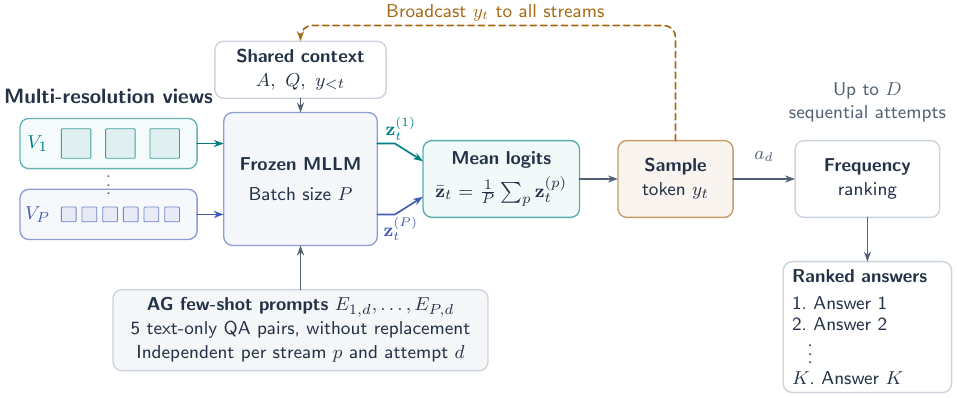}
    \caption{\textbf{Multi-resolution parallel decoding for audio-visual QA.}
    A frozen MLLM jointly processes $P$ spatio-temporal views with shared audio,
    and question.
    Accepted answers from up to $D$ sequential attempts are ranked by frequency.}
    \label{fig:architecture}
\end{figure}

To address this limitation, we propose scaling Video LLMs in parallel while varying the spatio-temporal resolution across the parallel streams.
Multi-rate and multi-resolution representations can capture complementary signals, such as global semantic information and local motion signals.
In addition, inspired by previous work on information sharing between parallel streams \cite{wang2026think,chen2026parallel,chung2026video}, we aggregate the logits at each decoding step across all streams.
This way, our approach can integrate information from diverse visual representations.
For the answer-generation task, we use few-shot prompting to adapt the model to the required format, using last year's training dataset (i.e., TREC 2025 VQA \cite{awad2025video}).
Finally, motivated by self-consistency decoding \cite{wang2022self}, we sample reasoning trajectories in successive generation attempts, each conditioned on $P$ parallel streams, and produce a frequency-based ranking of the accepted answers.
The complete architecture is shown in Figure~\ref{fig:architecture}.

We evaluate our proposed approach in the context of the TREC 2026 VQA track \cite{awad2026video}, on both multiple-choice (MC) and open-ended answer generation (AG) video QA tasks.
The TREC VQA track provides a common benchmark for assessing systems, allowing us to examine the effectiveness of our approach in a standardized setting.
This year, the VQA evaluation dataset contains a mix of YouTube short-form videos and a subset of longer scenes from a licensed movie dataset, the latter with an average duration of about 1--2 minutes.
Furthermore, it includes both human and automatically generated QA annotations.
Importantly, the queries may require reasoning across modalities, including audio, video and text.
In the remainder of this paper, we describe the related literature, our approach, our submitted runs and we analyze their results on the TREC evaluation.

\section{Related Work}

Parallel scaling increases the number of parallel generations of a large language model (LLM) at inference time.
The foundational work in this domain is self-consistency decoding \cite{wang2022self}, which samples diverse reasoning trajectories and aggregates their answers by majority voting.
Later work selects among repeated samples using verifiers, majority voting or reward models \cite{brown2024large}, or trains a compact model to aggregate the sampled answers \cite{qi2025learning}.
More recent methods combine the parallel streams during decoding instead of aggregating final answers.
For instance, ParScale \cite{chen2026parallel} trains the model with $P$ streams using learned prefixes and a learned per-token aggregation of their outputs, whereas the training-free ThinkMerge \cite{wang2026think} averages next-token logits across independent reasoning traces while decoding the answer.

In the video domain, Goulas et al. \cite{goulasmllm} rank ten completions, each generated from a random frame subset, by frequency.
Suo et al. \cite{suo2025trial} select among completions using a self-reward score, with an external LLM categorizing the question.
VidCtx \cite{goulas2025vidctx} answers multiple-choice questions by processing each frame separately and pooling the answer probabilities across frames.
Most similar to our approach, Video Parallel Scaling \cite{chung2026video} assigns each stream a frame subset by sampling with a fixed offset and aggregates the output probabilities of the streams.
In contrast, we vary the spatio-temporal resolution across streams to capture complementary signals, and we address Video QA under Chain-of-Thought reasoning instead of direct generation.

Combining visual representations sampled at different rates was popularized by SlowFast \cite{feichtenhofer2019slowfast}, which pairs a fast high-rate pathway with a slow low-rate pathway, and was later extended to video LLMs by LITA \cite{huang2024lita} and SlowFast-LLaVA \cite{xu2024slowfast}.
Instead of merging slow and fast representations into a single computation stream, we process them as parallel streams of the video LLM.

\section{Method}

Given a video $V$, question $Q$ and audio track $A$, we process one question at a time by batching $P$ parallel streams in a single expanded batch.
For a video with $N$ frames and duration $T$, stream $p$ has a sampling-rate budget $f_p$ and a frame cap $M_p$.
We select
\begin{equation}
n_p = \left\lfloor \min\left(N,M_p,f_pT\right) \right\rfloor
\end{equation}
frames using standard uniform frame sampling, spanning the entire video.
The resulting views $V_1,\ldots,V_P$ are constructed once per question and reused across generation attempts.
Each view has the same maximum visual-token budget $L$, applied separately to each stream.
We downscale views that exceed the corresponding pixel budget using bilinear interpolation, approximately preserving the aspect ratio.
The question $Q$, available audio track $A$, and generated prefix are shared across streams.
AG streams additionally receive independently sampled few-shot examples, as described below.
We use a frozen audio-visual Video LLM to perform batched decoding.
Inspired by prior work \cite{wang2026think,chen2026parallel,chung2026video}, we aggregate the logits across the $P$ parallel streams via mean pooling:
\begin{equation}
\bar{\mathbf{z}}_t = \frac{1}{P} \sum_{p=1}^{P} \mathbf{z}_t^{(p)}
\end{equation}
where $\mathbf{z}_t^{(p)}$ denotes the logits of stream $p$ at decoding step $t$, and $\bar{\mathbf{z}}_t$ denotes their mean.
Unlike ThinkMerge \cite{wang2026think}, which averages logits only while decoding the answer after independent reasoning traces, we pool the logits at every decoding step, including the reasoning tokens.
We apply temperature scaling and top-k/top-p filtering to the pooled logits and sample the next token $y_t$.
This token is then broadcast across the expanded batch, so all streams share the same generated prefix throughout an attempt.

Inspired by self-consistency decoding \cite{wang2022self}, we perform up to $D$ sequential generation attempts, each using the $P$ streams described above.
We stop this stage when we reach the target number of distinct accepted answers (i.e., 10 for AG, or the number of provided options for MC), or exhaust the attempt budget.
For AG, we group answers after collapsing whitespace and applying case folding, following the official validation code.
For MC, we group responses by the extracted option index.
Groups are ranked by decreasing frequency, with ties resolved by first occurrence.
For each AG group, we select one of its original answer strings uniformly at random for submission.

For AG, if we could not generate the required number of answers, we use the same frozen model to repeatedly paraphrase the highest-frequency original answer with a fixed text-only prompt containing the question and that answer.
The prompt instructs the model to produce a substantial wording change without changing the meaning of the answer.
For MC, we randomly shuffle the options that were not selected, to produce a complete answer list.

Finally, we use few-shot prompting to adapt the output style for AG without training.
Specifically, for every stream $p$ and generation attempt $d$, we sample a set $E_{p,d}$ of question/answer text examples from the 500 examples of last year's training dataset, without replacement.
The examples contain no associated video or audio.
Selections are independent across streams and attempts and may overlap.
This introduces prompt diversity across both streams and attempts, including for the $P=1$ AG baselines.
MC uses no few-shot examples.

\section{Experiments}

\subsection{Implementation Details}
We use the pre-trained \texttt{Qwen3-Omni-30B-A3B-Thinking} \cite{xu2025qwen3} model as the frozen Video LLM backbone, which is a Mixture-of-Experts (MoE) model that supports audio-visual inputs and is equipped with Chain-of-Thought (CoT) reasoning.
This variant consists of the Thinker component only and produces text output.
We load audio tracks at 16 kHz.
If audio loading fails, we use video-only input.
We set the maximum visual-token budget to $L=8192$ per stream, corresponding to a pixel budget of $8192\times 32\times 32\times 2$.
Generation uses temperature $1.0$, top-p $0.95$, top-k $20$ and a limit of $1024$ new tokens per call, including reasoning tokens.
We accept a response only if it contains the closing reasoning marker \texttt{</think>}.
For AG, each $E_{p,d}$ contains 5 examples.
All MC runs are zero-shot.
The prompts are listed in Appendix~\ref{sec:prompts}.
We run inference in \texttt{bfloat16} precision with FlashAttention-2.
The hardware and software environment is listed in Appendix~\ref{sec:environment}.
For missing or broken video files, we substitute a placeholder sequence of empty frames without an audio track.
The official MC results exclude 15 of 2,443 queries, leaving 2,428 evaluated queries.

\subsection{Metrics}
We present the following metrics, as reported by the TREC 2026 workshop organizers:
\begin{itemize}
\item For the MC task, we present the official top-1 accuracy and mean reciprocal rank (MRR).
Top-1 is used to evaluate the correctness of our approach, whereas MRR is used to evaluate the ranking quality of the system.
\item For the AG task, we present the official METEOR \cite{banerjee2005meteor}, BERTScore F1 \cite{zhang2020bertscore}, and semantic textual similarity (STS) scores.
We report the scores of the top-1 ranked answer, which measure the quality of the system's first prediction, and normalized discounted cumulative gain (NDCG) across the top-10 ranked answers, which measures the ranking quality of the submitted response set.
\end{itemize}
MC accuracy and MRR are averages over the 2,428 evaluated queries.
For AG, the Human Answers Only setting uses human reference answers.
Human + Alternative Answers augments this set with additional evaluation references.
Both settings evaluate the same submitted answers.

\subsection{Main Results}

We submitted 5 MC runs and 4 AG runs:
\begin{itemize}
\item For the MC task, we evaluate baseline CoT generation using two different fixed spatio-temporal resolutions, and we compare it with $P=2$ parallel generation using the combination of the two resolutions.
All MC runs are zero-shot.
To study the effect of the generation attempt budget $D$, we evaluate our main configuration with $D\in\{1,5,10\}$.
\item For the AG task, we compare single-stream CoT baselines at two fixed spatio-temporal resolutions with $P=2$ parallel generation combining both resolutions.
Note that the 4-fps baseline uses a 256-frame cap, whereas the 4-fps view in the two-stream run uses 128.
To study ranking, the \texttt{certhiti1x\_random} variant shuffles the ranked answer lists from \texttt{certhiti1x}, while keeping the answer strings unchanged.
All AG inference runs use 5 few-shot examples per stream and attempt, with an initial attempt budget $D=30$ and additional paraphrasing calls when needed.
\end{itemize}

In Tables~\ref{tab:mc-results} and~\ref{tab:ag-results}, $P$ denotes the number of parallel streams and $D$ the generation attempt budget.
Video views are specified as sampling-rate budget (fps) / maximum frames.

\begin{table}[htb]
    \centering
    \caption{\textbf{TREC 2026 VQA multiple-choice results.} Accuracy is reported as a percentage and MRR on the 0--1 scale. Higher scores are better. Best results are bold and second-best results are underlined.}
    \label{tab:mc-results}
    \begin{tabular}{@{}lcclcc@{}}
        \toprule
        Run ID & $P$ & $D$ & Views & Accuracy (\%) & MRR \\
        \midrule
        \texttt{certhiti1x\_mc}         & 1 & 5 & 4/128 & 80.7 & 0.867 \\
        \texttt{certhiti1x\_mc\_16fps}  & 1 & 5 & 16/256 & 80.4 & 0.868 \\
        \midrule
        \texttt{certhiti2x\_mc\_1depth} & 2 & 1 & 4/128, 16/256 & 80.4 & 0.855 \\
        \texttt{certhiti2x\_mc} (main)  & 2 & 5 & 4/128, 16/256 & \textbf{81.3} & \underline{0.871} \\
        \texttt{certhiti2x\_mc\_10dep}  & 2 & 10 & 4/128, 16/256 & \underline{81.1} & \textbf{0.874} \\
        \bottomrule
    \end{tabular}
\end{table}

At $D=5$, combining the two video views improves MC accuracy by 0.6 percentage points over the strongest single-stream baseline (Table~\ref{tab:mc-results}).
The main run reaches 81.3\% accuracy, compared with 80.7\% and 80.4\% for the 4-fps and 16-fps baselines, respectively.
Its MRR is 0.871, compared with the best baseline value of 0.868.

For the two-stream configuration, MRR increases from 0.855 to 0.871 and 0.874 as the attempt budget rises from $D=1$ to 5 and 10.
Top-1 accuracy is highest at $D=5$ (81.3\%), compared with 81.1\% at $D=10$.
Thus, $D=5$ achieves the best observed top-1 result with a smaller attempt budget than $D=10$, while $D=10$ gives the best MRR.

\begin{table}[htb]
    \centering
    \caption{\textbf{TREC 2026 VQA answer-generation results.} All scores are multiplied by 100. Higher scores are better. Within each reference setting, best results are bold and second-best results are underlined. Human-reference top-1 emphasis uses unrounded scores.}
    \label{tab:ag-results}
    \begin{tabular}{@{}lclcccccc@{}}
        \toprule
        & & & \multicolumn{3}{c}{Top-1 answer} & \multicolumn{3}{c}{NDCG@10} \\
        \cmidrule(lr){4-6}\cmidrule(l){7-9}
        Run ID & $P$ & Views & STS & METEOR & \shortstack{BERTScore\\F1} & STS & METEOR & \shortstack{BERTScore\\F1} \\
        \midrule
        \multicolumn{9}{@{}l}{\emph{Human Answers Only}} \\
        \texttt{certhiti1x}         & 1 & 4/256 & \underline{50.5} & \underline{23.4} & \underline{89.3} & \underline{89.9} & 77.2 & \underline{99.1} \\
        \texttt{certhiti1x16fps}  & 1 & 16/256 & 50.1 & 23.4 & 89.3 & \underline{89.9} & \underline{77.6} & \textbf{99.2} \\
        \texttt{certhiti1x\_random} & 1 & 4/256 & 44.7 & 22.3 & 88.1 & 88.0 & 76.9 & 98.9 \\
        \texttt{certhiti2x} (main)  & 2 & \shortstack[l]{4/128,\\16/256} & \textbf{50.7} & \textbf{24.5} & \textbf{89.5} & \textbf{90.2} & \textbf{78.0} & \textbf{99.2} \\
        \midrule
        \multicolumn{9}{@{}l}{\emph{Human + Alternative Answers}} \\
        \texttt{certhiti1x} & 1 & 4/256 & \underline{52.7} & \underline{25.6} & \underline{89.8} & \underline{90.2} & 78.4 & \underline{99.1} \\
        \texttt{certhiti1x16fps} & 1 & 16/256 & 52.3 & \underline{25.6} & 89.7 & \underline{90.2} & \underline{78.9} & \textbf{99.2} \\
        \texttt{certhiti1x\_random} & 1 & 4/256 & 46.9 & 24.8 & 88.6 & 88.5 & 78.2 & 98.9 \\
        \texttt{certhiti2x} (main) & 2 & \shortstack[l]{4/128,\\16/256} & \textbf{53.0} & \textbf{26.9} & \textbf{89.9} & \textbf{90.5} & \textbf{79.2} & \textbf{99.2} \\
        \bottomrule
    \end{tabular}
\end{table}

\begin{figure}[tb]
    \centering
    \includegraphics[width=\linewidth]{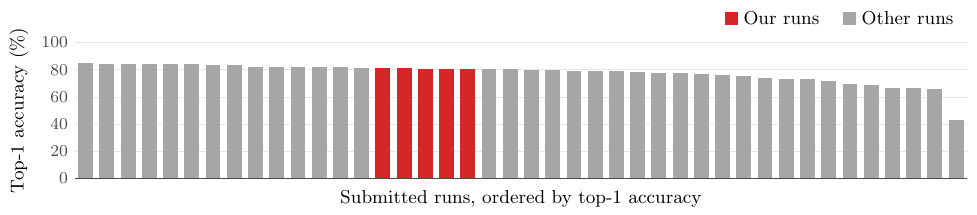}
    \caption{\textbf{MC results across all TREC 2026 VQA participants.}
    Each bar is one submitted run, ordered by top-1 accuracy (multiplied by 100).
    Our runs are shown in red.}
    \label{fig:mc-comparison}

    \vspace{6pt}
    \includegraphics{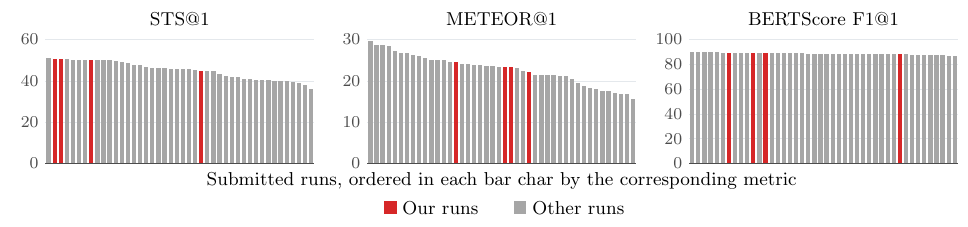}
\caption{\textbf{AG results across all TREC 2026 VQA participants.}
Each bar is one submitted run, and each bar chart is ordered by its own metric (multiplied by 100).
Our runs are shown in red.
}
\label{fig:ag-comparison}
\end{figure}

Figure~\ref{fig:mc-comparison} and Figure~\ref{fig:ag-comparison} place our MC and AG results among all submitted runs in the TREC 2026 VQA task.
For MC, our main run reaches 81.30\%, compared with 84.76\% for the highest-scoring submission.
For AG, our main run ranks second among the 44 submitted runs by STS@1.
However, note that the results between participants are not strictly comparable, since other runs may use different Video LLM backbones and inference setups, including proprietary models with significantly higher number of parameters.

The two-stream run exceeds both single-stream baselines on all three human-reference top-1 AG metrics (Table~\ref{tab:ag-results}).
Relative to the 4-fps baseline, the differences are 0.2 STS@1, 1.1 METEOR@1 and 0.2 BERTScore F1@1 points.
The 4-fps baseline exceeds its submitted shuffled variant by 5.7 STS@1, 1.1 METEOR@1 and 1.2 BERTScore F1@1 points, supporting frequency-based ordering.

\subsection{Runtime and Generation Time}

\begin{table}[ht]
\centering
\caption{\textbf{Runtime and generation time.} Approximate total runtime and mean recorded generation time per submitted AG answer on a single NVIDIA RTX PRO 6000 Blackwell GPU.}
\label{tab:latency}
\begin{tabular}{@{}lcccc@{}}
\toprule
& \multicolumn{2}{c}{Multiple-choice} & \multicolumn{2}{c}{Answer-generation} \\
\midrule
Parallel streams $P$ & 1 & 2 & 1 & 2 \\
Generation attempt budget $D$ & 5 & 5 & 30 & 30 \\
\midrule
Total runtime (h) & 29 & 34 & 159 & 180 \\
Mean generation time per submitted answer (s) & N/A & N/A & 13.0 & 16.6 \\
\bottomrule
\end{tabular}
\end{table}

Using two streams increases total runtime by approximately 17\% for MC and 13\% for AG in the measured configurations (Table~\ref{tab:latency}).
Runtime rises from 29 to 34 hours for MC and from 159 to 180 hours for AG.
The mean recorded generation time per submitted AG answer increases from 13.0 to 16.6 seconds (approximately 28\%).

Total runtime covers a complete evaluation run, including initialization, data loading, pre-processing, generation, parsing, ranking and paraphrasing.
These approximate durations also include overhead related to checkpointing/resuming and networked-filesystem transfers.
The per-answer measurement averages the recorded generation times attached to the submitted AG answers.
All evaluations process one question at a time on a single RTX PRO 6000 Blackwell GPU, with an expanded batch of $P$ streams.
The observed scaling behavior is consistent with prior findings: data movement has been identified as the key bottleneck in Transformer training \cite{ivanov2021data}, and utoregressive decoding at small batch sizes is largely memory-bound rather than compute-bound \cite{chen2026parallel}.
Therefore, batched execution can amortize access to the model weights across streams, offering a latency-friendly way to scale inference compute.

\section{Conclusion}

We presented a training-free approach to video question answering that scales a video LLM in parallel, integrating multiple spatio-temporal views with varying resolutions.
Each generation attempt processes the views jointly, while averaging the logits across the parallel stream axis at each decoding step.
Successive attempts provide answers for frequency-based ranking, with task-specific completion of the candidate lists.
At $D=5$, the two-stream MC run exceeds the strongest single-stream baseline by 0.6 percentage points in top-1 accuracy.
The two-stream AG run outperforms both baselines on the three human-reference top-1 metrics.
Finally, the two-stream parallel runs are shown to have a low to modest latency overhead in practice, compared to single-stream generation.

\section*{Acknowledgements}

This work was supported by project eXeLMM, implemented in the framework of H.F.R.I. call ``3rd Call for H.F.R.I.'s Research Projects to Support Faculty Members \& Researchers'' (H.F.R.I. Project Number: 25957).



\appendix

\section{Prompts}
\label{sec:prompts}

\medskip
\noindent
\begin{minipage}[t]{0.48\linewidth}
\promptbox{Multiple-choice}{%
$\langle$video$\rangle$\\
$\langle$question$\rangle$\\
Options:\\
A. $\langle$option 1$\rangle$\\
B. $\langle$option 2$\rangle$\\
\dots\\
Return only the correct option's letter.}

\medskip
\promptbox{Paraphrasing}{%
Rephrase the following answer to the given question. Return only the rephrased answer. The rephrased answer should differ significantly from the original answer, without altering its meaning.\\
Question: $\langle$question$\rangle$\\
Original Answer: $\langle$highest-frequency answer$\rangle$}
\end{minipage}\hfill
\begin{minipage}[t]{0.48\linewidth}
\promptbox{Answer generation}{%
Answer the video question in the same style as the examples.\\[0.5\baselineskip]
\#\#\# Examples\\[0.5\baselineskip]
Question: $\langle$example question 1$\rangle$\\
Answer: $\langle$example answer 1$\rangle$\\[0.5\baselineskip]
\dots\\[0.5\baselineskip]
Question: $\langle$example question 5$\rangle$\\
Answer: $\langle$example answer 5$\rangle$\\[0.5\baselineskip]
\#\#\# End of examples\\[0.5\baselineskip]
\#\#\# Task\\[0.5\baselineskip]
$\langle$video$\rangle$\\
$\langle$question$\rangle$}
\end{minipage}

\section{Environment}
\label{sec:environment}

Table~\ref{tab:environment} lists the hardware and software environment of our experiments.

\begin{table}[h!]
    \centering
    \caption{\textbf{Hardware and software environment.}}
    \label{tab:environment}
    \begin{tabular}{@{}llll@{}}
        \toprule
        \multicolumn{2}{@{}l}{Hardware and system} & \multicolumn{2}{l@{}}{Software} \\
        \midrule
        GPU & NVIDIA RTX PRO 6000 Blackwell (96\,GB) & Python & 3.12.3 \\
        System memory & 250\,GB & PyTorch & 2.10 \\
        Operating system & Ubuntu 24.04 LTS & Transformers & 5.16.1 \\
        CUDA & 12.8 & & \\
        \bottomrule
    \end{tabular}
\end{table}

\end{document}